# Learning Automata of PLCs in Production Lines Using LSTM

[1]Iyas AlTalafha, [2] Yaprak Yalçın, [3] Gülcihan Özdemir
[1][2][3] *Istanbul Technical University*
[1]altalafha20@itu.edu.tr,[2] yalciny@itu.edu.tr, [3]ozdemirg@itu.edu.tr

*Abstract— Production Lines and Conveying Systems are the staple of modern manufacturing processes. Manufacturing efficiency is directly related to the efficiency of the means of production and conveying. Modelling in the industrial context has always been a challenge due to the complexity that comes along with modern manufacturing standards. Long Short-Term Memory is a pattern recognition Recurrent Neural Network, that is utilised on a simple pneumatic conveying system which transports a wooden block around the system. Recurrent Neural Networks (RNNs) capture temporal dependencies through feedback loops, while Long Short-Term Memory (LSTM) networks enhance this capability by using gated mechanisms to effectively learn long-term dependencies. Conveying systems, representing a major component of production lines, are chosen as the target to model to present an approach applicable in large scale production lines in a simpler format. In this paper data from sensors are used to train the LSTM in order to output an Automaton that models the conveying system. The automaton obtained from the proposed LSTM approach is compared with the automaton obtained from OTALA. The resultant LSTM automaton proves to be a more accurate representation of the conveying system, unlike the one obtained from OTALA.*

## I. INTRODUCTION

Manufacturing processes, as technology advances, have become more efficient and in turn more complex. The industrial evolvement allowed for the inclusion of many apparatus that improved upon the production output in order to satisfy the balance of supply and demand. Therefore, the importance of monitoring, controlling and maintaining the production lines has increased and became of equal focus to manufacturers as the manufacturing process itself. The operational state of the production lines is proportional to the quality, speed and quantity of output and to meeting the production quota, which in turn holds financial and competitive significance.

However, the issue arises when manufacturing processes utilize several components for operation and thus resulting in a complicated system. Employing a collection of methods to maintain the desired output and operation of the system, ensures minimizing possibility of fault occurrence and product rejection rates, additionally it boosts the safety of operators.

Programmable Logic Controllers (PLCs) are considered part of the foundation of any manufacturing plant. Especially since they have been developed to monitor and control production processes with the purpose of ensuring operation within expected parameters by providing comprehensible feedback and a digitized visualization of the system processes. Since PLCs act as the intermediary interface between the manufacturing line and the operator, data from sensors, monitored and managed by PLCs, can be acquired to model the entire manufacturing process. Modelling allows for the readability of the production process and does not necessarily rely on the mathematical representation of individual machinery. Relying on the PLC inputs and outputs to model the system focuses on the entire network or individual components, thus allowing for freedom in modelling.

In this paper, Long Short-Term Memory is used to create an automaton model of a rudimentary pneumatic conveying system, and is compared with the OTALA algorithm. In section (2) a theoretical background is established, where the definition of an Automaton, the learning of Automata and theory of the LSTM structure are covered. In section (3) the state of the art is discussed, in which the nature of the data collected and the pneumatic conveying system are described. In section (4) the obtained results are discussed, in addition to the algorithms utilized. Finally, in section (5) a conclusion and a possible expansion on the proposed approach is briefly discussed.

## II. THEORETICAL BACKGROUND

### A. Automata

In a random environment, where only a finite set of actions is available, each action triggers a response that may be either favourable or unfavourable. The challenge in designing an automaton lies in determining how to choose the next action based on past actions and responses, despite having only limited knowledge about the environment's underlying nature. An Automaton can be either deterministic or stochastic. [3] The stochastic automaton continuously updates the probabilities of its various actions as new information appears.

In the context of automata theory, an automaton is characterized by a quintuple that includes a set of internal states, a set of input actions, a set of outputs, a state-transition function mapping the current state and input to the next state, and an output function mapping the current state and input to the current output. [2] If the output depends solely on the current state, the automaton is known as a state–output automaton. [3] Both models are considered finite when their sets of states, inputs, and outputs are finite. The automaton is deterministic if both the state-transition and output functions yield unique results for a given initial state and input; if either function is stochastic, then the automaton is referred to as stochastic, meaning its behavior can only be described in terms of probabilities for subsequent states and actions.

In the context of this paper, the resulting automaton is a deterministic state-output automaton since the transitions between states do not rely on probability measures and each observation corresponds to a unique state. Additionally, the observations recorded are sequential and the order of occurrence determines the transitions from one state to the next.

### B. Learning Automata

A learning automaton is a self-adaptive decision-making model that interacts with its environment to optimize a specific objective. In the context of production lines, learning automata can model the behavior of programmable logic controllers (PLCs) and their interaction with the system's sensors and actuators. The automaton learns the optimal sequence of states and transitions that represent the production line's operations by processing binary input data, such as sensor readings. An Automaton can be defined by the Input signals $(I)$, or the observations, the Output Signals $(O)$, referred to as the State Values in the OTALA section, the set of finite States $(S_n)$, the mapping function $(f)$ and the Transitions $(T)$ between the states. [2]

$$A = (I, O, S_n, f, T) \quad (1)$$

In the context of this paper, the learning of the automata in order to visualise a model of the conveying system, LSTM, an RNN learning methodology considered ideal for sequential observations was employed. In order to measure efficiency, it has been compared with OTALA, an automaton learning algorithm.

### C. Long Short-Term Memory

LSTM stands for **Long Short-Term Memory**. It is a type of recurrent neural network (RNN) designed to learn long-term dependencies in sequential data. Traditional RNNs suffer from problems like vanishing or exploding gradients when the sequences are long, making it hard for them to capture long-term dependencies. LSTM overcomes these issues by using a special architecture that includes **memory cells** and several **gating mechanisms** to control the flow of information. [8] Due to the sequential nature of the data LSTM is a fitting learning method. The order of the observations allows the recognition of the movement of the wooden block around the conveyor system and the transitions between each state.

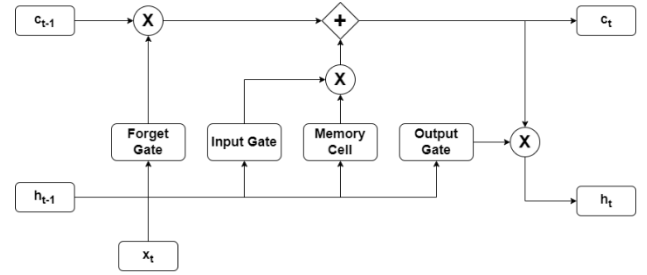

**Figure 1: Long Short-Term Memory Structure. [8]**

### III. STATE OF THE ART

#### A. Conveyor Systems and Production Lines

The conveying system, seen in **Fig. 2**, from which the data was collected in order to model an automaton, is controlled via a Siemens PLC and operates pneumatically to move a wooden block around the system in a looping manner. For better referencing and visualization, each corner was named in the order at which the block travels. The block starts at **Position A**, it is then picked up by the suction arm and moved to **Position B**, where it is placed on the next platform. From **Position B** it is moved on to the next Platform in **Position C**, after which it is moved onto the next platform in **Position D**. Finally, the last platform moves to **Position D**, where the block slides onto the platform and it is transferred back to **Position A**, finishing a singular conveying cycle.

The system is operated using three buttons that starts and stops the system operation and also activates the pneumatic compressor that drives the platforms on the conveying system. There are eleven proximity sensors that detect where the platform responsible for moving the block is. By relying on the sensor readings and how they change in value, the order at which the block moves can be determined. Initially, the sensor readings at the moment of starting the conveying system are recorded. Then at each time step the sensors are read again. Each sensor reading corresponds to a state in which the system can be defined with.

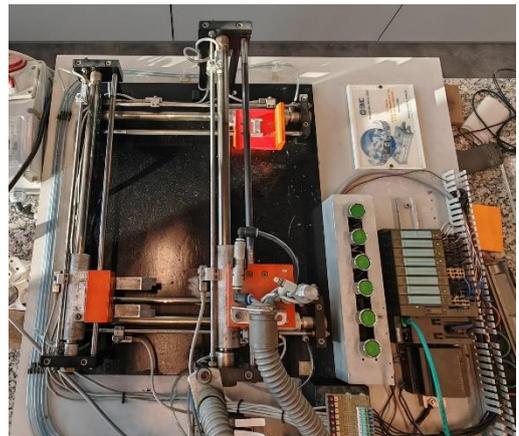

**Figure 2: Pneumatic Conveying System in Industrial Automation Laboratory in Istanbul Technical University.**

The image above is of the pneumatic conveyor system employed in the exploration of the LSTM modelling approach. As is visible the suction Arm is above Position B amid the transitioning of the wooden block from the starting position, Position A. The cyclic nature of the data is evident in the looping architecture observed in the image. Where the path the block travels can be described as a closed looping path about four corners, referred to as positions, with platforms labelled to ease the referencing to the system and state definition for the observations.

Sensors present on the conveying system are clarified in the diagram of the conveyor system present in **Fig. 3.**

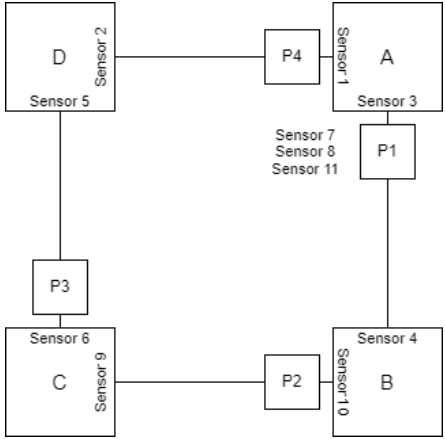

**Figure 3: Conveyor System Diagram.**

**Table: Expected States for normal system operation.**

| States | System Status |
|---|---|
| 1 | START |
| 2 | Arm (**P1**) going **Down**/Suction **OFF** |
| 3 | Arm (**P1**) **Down**/Suction **ON** |
| 4 | Arm (**P1**) Picking Block **Up** |
| 5 | Block is fully **Up**/Suction **ON** |
| 6 | Block is held by Arm/Arm is moving towards **Position B** |
| 7 | Arm (**P1**) is holding Block on **Position B**/Suction **ON**/ Arm is **Up** |
| 8 | Arm placing Block **Down** on **P2**/Suction **ON** |
| 9 | Block is on **P2** on Position B/Suction **OFF**/Arm is **Down** |
| 10 | Arm is going back **Up** / Suction is **OFF** |
| 11 | Arm is fully back **Up**/Block is on **P2** |
| 12 | **P2** is carrying Block towards **Position C/P1** is going back to **Position A** |
| 13 | Block is on **P2** on **Position C/P1** back to **Position A** |
| 14 | Block is on **P3** on **Position C/P2** is between **B** and **C** |
| 15 | **P3** is moving towards **Position D/P2** is between **B** and **C** |
| 16 | Block on **P3** on **Position D/P4** between **D** and **A/P2** back to **Position B** |
| 17 | Block is on **P4** on **Position D** |
| 18 | **P4** is moving towards **Position A** |
| 19 | Block on **Position A** on **P4/P3** between **A** and **C/Arm** (**P1**) is going down |
| 20 | END (Arm starts again) |

The above table, describes the normal expected state-wise transition and operation of the conveyor system. The states were obtained from observing the system operation and recording the order of events that take place to complete a single cycle, which was obtained to have as a reference to the entire system operation. Where P1 refers to the first platform which is the pneumatic Arm. P2 is the second platform which traverses between Positions B and C. The Third platform moves about Positions C and D and finally the Fourth platform moves between D and A.

*B. Nature of the Data Collected*

It is important to note that the conveying system continues moving the block cyclically and the cycle does not terminate after the loop ends, unless it was terminated by the operator. The cyclic nature of the conveying system is crucial to the modelling logic applied in this paper, where it represents the continuous operation of machinery in real life manufacturing processes. Where the stoppage of machines could delay production and manufacturing processes, especially if initial boot-up costs were considered, in addition to the necessity of the continuous operation of production lines in order to maintain the manufacturing quota.

The input used was collected via PLC with a sampling rate of 500 milliseconds for a total duration of 10 minutes resulting in 1200 observations. The input array is a 1200 x 11 sized array, where each column is a binary value corresponding to a sensor value. A 1 signifies the platform is present at the sensor and a 0 signifies there is no platform detected. For the applications of LSTM and OTALA one cycle is chosen at random from the observations collected. In addition, the sensor readings have been recorded for when the block is at each of the four positions in order to pinpoint the sequential order of the transitions between the states. The movement of the wooden block around the pneumatic conveyor system is smooth and therefore does not require a large number of layers in the LSTM architecture.

### IV. MAIN RESULTS

*A. Automata creation with proposed LSTM*

The proposed LSTM approach can be described as a sequence to sequence classification. Where the emphasis is on the significance of the sequential nature of the cyclic data. The data is first divided into training and testing arrays, where 80% of the data is used to train the model and 20% is used for testing. The training data is comprised of forty cycles each cycle having varying lengths. The testing data is comprised of eleven cycles, each having varying lengths as well. The LSTM Network uses eleven features, corresponding to the number of the sensors in the system and five classes, for the five unique classes of the system, referring to the positions and transitions of the block being conveyed (A, B, C, D, Transition).

The LSTM includes fifty hidden layers, where it was determined to be the most accurate after attempting the training while varying the number of hidden layers. Additionally, softmax is used for sequence prediction, as mentioned previously in this section. In the context of classification an extended version of cross-entropy, **crossentropyex** is used, where the loss function is built to calculate the cross-entropy loss for each time step and then aggregates by averaging the losses over the entire sequence. This type of cross-entropy is used instead of the standard, due to the necessity to handle multiple outputs per sequence, which is the case of the data used, where some of the raw observations could be repeated and refer to different states.

The **Adaptive Moment Estimation (adam)** optimizer updates the network weights using the Adam optimization algorithm, where the learning rates are adapted during training leading to faster and more stable convergence, as can be seen in the Accuracy and Loss figure in **Fig. 7** below. The optimizer computes individual adaptive learning rates for each parameter using estimates of the first moment (the mean) and the second moment (the uncentered variance) of the gradients. It combines the advantages of two other extensions of stochastic gradient descent—namely, Adaptive Gradient Algorithm (AdaGrad) and Root Mean Square Propagation (RMSProp). This makes it both computationally efficient and well-suited for problems with large datasets and high-dimensional parameter spaces. Finally, the network is trained using the above parameters and the predicted states and transitions (automaton) is generated by relying on the Testing data, as discussed earlier in this section. [13]
Please see the flowchart of proposed LSTM method for automata creation in **Fig. 4** and see the pseudocode of the algorithm in the Appendix.

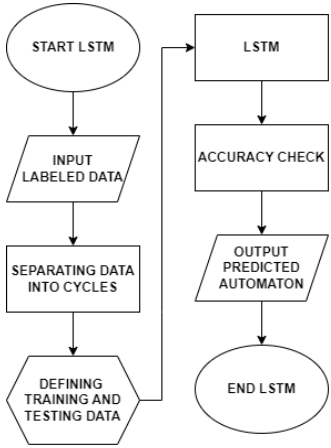

**Figure 4: Flowchart of proposed LSTM method for automaton creation.**

### B. Application of OTALA

The algorithm below applies the **OTALA** [2] algorithm in the context of the conveying system described for one cycle of operation.

**OTALA Algorithm:**
    **Given:** Observations $D_n$, Block Positions $P_n$
    **Result:** Automaton $A$
        For all observations $D_n$:
1. If $D_n \neq D_{n+1}$
$$d_n = D_n$$
2. If $P_n = f(d_n)$
$$S'_n = d_n$$

    else
$$S'_n = CreateNewState()$$
3. $T = T \cup \{(S'_{n-1}, S'_n)\}$
4. *Output* **Automaton** $A$

The inputs to the algorithm are the observations collected from the conveyor system sensor readings at a sampling rate predetermined in the PLC controlling the system and position references $P_n$. The resulting observations $D_n$ included repeated readings when the system is idle or is moving slower than the sampling rate. This prompted the inclusion of step **(1)** in the above algorithm, where it omits any repeated observations in order to maintain the integrity of the used Data, resulting in $d_n$. Step **(2)** refers to the sensor readings when the block being moved is at either of the four positions, where each observation in the data set obtained in **(1)** is checked to match the block positions defined by $P_n$ using a mapping function $f$, otherwise it creates a new state. Step **(3)** records the transitions $T$ between each state and the **Automaton** is outputted in the final step.

The learning of the automaton is considered complete when the observations match the final position of the block in the conveying loop, where the block returns to the starting position A. The learning of the Automaton can be measured according to the Number of States, the Transitions between the states and the resulting State Values. The number of states will correspond to a concise description of the movement of the block about the positions in the conveying system. The transitions recorded also ensure the sequential order of the states, in addition to providing valuable information about the position of the block in the system. Furthermore, the values of each unique state created further assures the completion of the loop, where the sensor readings match the expected sensor observations for one cycle.

### C. Results and Evaluation

The automaton in **Fig. 5** describes the transition of states resulting from OTALA after inputting observations for a single cycle, where each set of sensor readings represents a state of the system. The input sensor observations for a single cycle were a collection of thirty sensor readings. The observations obtained were divided into approximately fifty-one cycles. Each cycle was used as an input to OTALA and the automaton representing the conveying system best was chosen. The automaton consists of twelve states. According to the automaton in **Fig. 5**, State (1) is the starting state of the system. States (2), (3) and (4) correspond to the movement of the pneumatic suction Arm to place the Block onto the next platform that moves between Positions B and C. In States (5) and (6) the Block is on Position B. State (7) and (8) the block is in transition to Position C. State (9) the block is on Position C. State (10) the block is in transition to Position D. State (11) the block is on Position D. Finally, in State (12) the Block is on Position A, thus completing the cycle. However, it is important to note that the transition states of the block from Position D to Position A is not found in the Automaton obtained from OTALA.

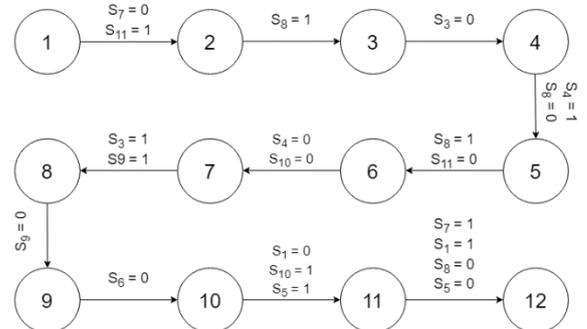

**Figure 5: Resultant OTALA Automaton.**

The diagram below is the automaton created using LSTM. This automaton defines the system more accurately, with more states. The model has an accuracy of 90% - 93% after training it over 1000 epochs. The differentiating characteristic of the automaton obtained using LSTM from that obtained using OTALA is that OTALA relies on a truth table determining exact positions corresponding to their observations, while the LSTM model is trained using a labeled data set and tested using an unlabeled data set. Furthermore, the trained model did not require as much intervention in the data as did the OTALA approach. Where both input datasets were shortened by omitting repeated observations, to minimize noise, resulting from the conveyor system staying idle, while the PLC recorded the sensor readings every 500 milliseconds.

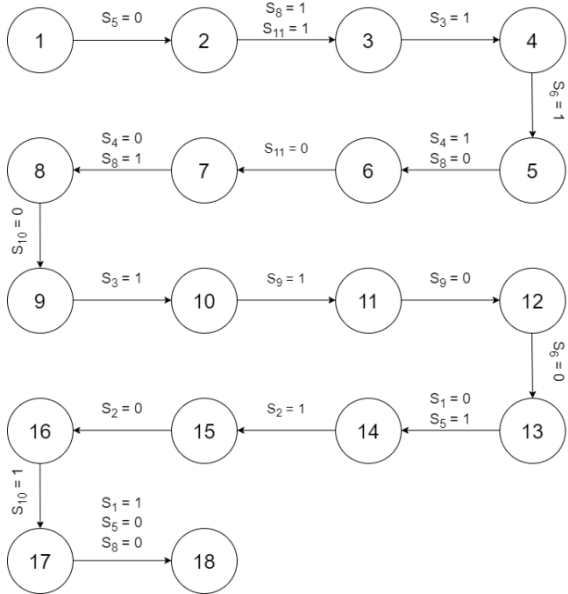

Figure 6: Resultant LSTM Automaton.

In **Fig. 6**, **State (1)** the Block is on **Position A**, the starting position of the system. **States (2) - (6)** describe the states Transition and the movement of the pneumatic suction arm towards **Position B**, on **State (7)**. **States (8) – (10)** the platform carrying the wooden block moves to **Position C**, on **State (11)**. **States (12)** and **(13)** the Platform moves towards **Position D**, on **State (14)**. Finally, during **States (15) – (17)** the block is moved towards **Position A**, on **State (18)**, thus finishing a cycle.

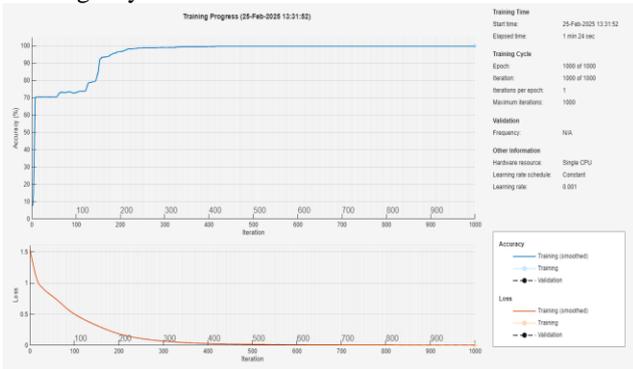

**Figure 7: LSTM Loss and Accuracy Convergence.**

The above graph shows the convergence of the Training and Testing of the LSTM model. It can be seen that after approximately five hundred iterations, Accuracy converges to approximately 99%. After repeated running of the algorithm the accuracy of the predicted states was maintained at a range of 90% to 94% in addition to a constant convergence after approximately 500 iterations.

## V. CONCLUSION AND OUTLOOK

The resultant automaton from the proposed LSTM approach, is more reliable when compared with the automaton from OTALA. Despite the fewer number of states in the automaton obtained through OTALA, it lacks a few important transitions and states essential for an accurate system description. The reliance on a mapping function in order to create a system accurate automaton complicates and reduces the applicability of OTALA on varying production lines. While the LSTM approach to model the system, does not depend on a system specific mapping function. Furthermore, the proposed sequence by sequence LSTM approach considers the sequential nature of production lines, where a certain order of processes must be followed in order to achieve the desired end product. Additionally, OTALA creates a singular automaton per production cycle and heavily depends on the data collection methodology, which increases uncertainty and the possibility of erroneous automaton creation. The proposed LSTM approach utilizes the sequence patterns of the data and accounts for sampling mishaps in the data collection portion. Future additions and advancements to this approach could be in the field of fault detection and classification, where the created automaton could be used to identify abnormal operation within a specific threshold.

## APPENDIX

**LSTM Pseudocode:**
   **Input:** Labeled Sensor Observations.
   **Output:** Automaton.
1. Read sensor data.
2. Identify indices where label "A" occurs:
   FOR each index i in labArr:
      IF labArr[i] == "A" THEN
         Append i to posA_id
3. Detect cycles:
   FOR j = 1 to length(posA_id)-1:
      cycleRange ← posA_id[j] to posA_id[j+1]
      Append cycleRange to cycles
4. For each cycle in cycles:
   Extract sensor data and labels.
   Store in cycD and cycL structures.
5. Prepare training and testing datasets:
   FOR first trainLength cycles:
      XTrain[c] ← Transpose(cycD[c].data)
      YTrain[c] ← Transpose(cycL[c].labs)
   FOR next testLength cycles:
      XTest[b]←Transpose(cycD[b+offset].data)
      YTest[b]←Transpose(cycL[b+offset].labs)
6. Convert YTrain and YTest to categorical type.
7. Validate training data:
   FOR each training cycle:
      IF number_timesteps in XTrain ≠ numLabels in YTrain THEN
         Report error
8. Define LSTM network architecture.
9. Set training options and train the network.
10. Classify test sequences using the trained network.
11. Calculate test accuracy by comparing predictions with YTest.